\documentclass[runningheads]{llncs}
\usepackage[T1]{fontenc}
\usepackage[utf8]{inputenc}
\usepackage{dsfont}
\usepackage{amsmath,graphicx,verbatim,amsfonts,booktabs, subcaption}
\usepackage[normalem]{ulem}   
\usepackage{xcolor}
\usepackage{textcomp} 
\newcommand{\indicator}{\mathbf{1}}
\begin{document}
\title{Contrastive uncertainty learning for robust multi-structure segmentation in lower-pelvic MR}
\titlerunning{Contrastive uncertainty learning}
\authorrunning{B. Sidiqi et al.}
%

\author{Bilal Sidiqi\inst{1,2} \and
Yipei Wang\inst{1,2}\orcidID{0000-0002-9589-7177}  \and
Wen Yan\inst{1,2}\orcidID{0000-0002-3962-5994} \and
Shaheer U. Saeed\inst{1,3,4}\orcidID{0009-0002-9889-9853} \and
Daniel C. Alexander\inst{1,5}\orcidID{0000-0003-2439-350X} \and
Yipeng Hu\inst{1,2}\orcidID{0000-0003-4902-0486} 
}

\institute{UCL Hawkes Institute, University College London, London, UK \\ \email{bilal.sidiqi.21@ucl.ac.uk} \and
Department of Medical Physics and Biomedical
Engineering, University College London, London, UK \\ \and
Centre for Bioengineering, School of Engineering and Materials Science, Queen
Mary University of London, London, United Kingdom \\ \and 
Digital Environment Research Institute, Queen Mary University of London,
London, United Kingdom \\ \and
Department of Computer Science, University College London, London, UK
}
  
\maketitle              
\begin{abstract}
Boundary delineation of anatomical structures in prostate MRI is often compromised by regional ambiguity. Automated segmentation methods relying on global overlap loss functions tend to be insensitive to surface deviations, leading to significant errors in ambiguous regions. Most existing uncertainty-aware methods treat uncertainty as a post-hoc quality indicator or prediction filter.
We propose Contrastive Uncertainty Learning (CUL), a supervised framework that uses predictive uncertainty to directly guide representation learning. Using Monte Carlo Dropout, CUL identifies voxel-wise uncertainty and structures the embedding geometry accordingly: low-uncertainty, correctly predicted features form stable voxels while high-uncertainty or misclassified features are actively repelled, preventing feature collapse at difficult boundaries.
We evaluate CUL on a 589-scan 3D prostate MRI multi-structure segmentation task (8 anatomical structures) using a ViT with UperNet decoder architecture. Average Dice ($83.38\%$) remains comparable to the baseline ($83.35\%$), while surface distance decreases significantly ($p<0.001$) across 6 of 8 anatomical structures. A hard-case analysis reveals a $22{\times}$ asymmetry between significantly improved and degraded cases ($18.8\%$ vs $0.9\%$). Ablation studies show that removing uncertainty-guided sample selection eliminates improvements for the transition zone, while replacing the contrastive objective with a boundary loss additionally removes bladder improvements, demonstrating that both uncertainty-guided sampling and the contrastive objective are essential for the observed gains. These findings demonstrate that the proposed uncertainty-guided representation learning is a useful mechanism for reducing the risk of extreme errors in safety-critical medical segmentation

\end{abstract}

\section{Introduction}
Deep learning based segmentation is becoming important for quantitative analysis of prostate MRI, enabling automated delineation of prostate zones and associated structures for prostate cancer diagnosis, risk stratification, and treatment planning \cite{turkbey2022deep}. Despite strong average performance, convolutional and transformer models often fail on boundary ambiguous cases, producing large boundary deviations that are not fully reflected by global overlap metrics \cite{Fassia2024Deep}. A key challenge lies in learning representations that remain robust in regions of high predictive uncertainty, typically near boundaries where standard losses such as Dice or cross-entropy do not explicitly incorporate confidence information.

Predictive uncertainty estimates represent model confidence and are widely used in medical image analysis. Bayesian approximations, including Monte Carlo dropout \cite{Gal2016} and deep ensembles \cite{Lakshminarayanan2017}, correlate uncertainty with segmentation error and support quality assurance, error detection, and selective prediction in safety-critical applications.

Uncertainty has also been incorporated into training strategies, particularly in semi-supervised segmentation. In teacher–student frameworks, uncertainty is used to filter unreliable pseudo-labels or weight consistency losses to mitigate confirmation bias \cite{Ouali2020}. Uncertainty guided contrastive learning similarly excludes high uncertainty pixels from positive pair construction to stabilise representations \cite{Wang2022UGPCL,Hua2022}. Prototype based and memory based approaches construct class prototypes from confident regions while preventing ambiguous voxels from contaminating cluster centres \cite{Chen2022CPSPrototype}. These methods improve robustness when labelled data are limited, using uncertainty as a filtering or weighting mechanism.

Directly using uncertainty to structure embedding geometry in fully supervised segmentation remains underexplored, particularly for intrinsically ambiguous multi-structure MRI segmentation.

Contrastive learning provides a principled mechanism for structuring representation space. Self-supervised frameworks such as MoCo \cite{He2020}, SimCLR \cite{Chen2020}, and supervised contrastive learning \cite{Khosla2020} demonstrate their ability to shape embeddings. Contrastive learning at the pixel and voxel levels has been adopted in segmentation to enhance intra-class compactness and inter-class separation. However, these methods typically define positive pairs solely by class identity or spatial proximity, without considering predictive confidence. 

We hypothesise that structuring embedding space based on predictive uncertainty can reduce catastrophic boundary deviations while preserving overall segmentation accuracy. Confident same-class voxels should anchor compact clusters, whereas high-uncertainty voxels should be prevented from distorting these regions of feature space.
We therefore propose Contrastive Uncertainty Learning (CUL), a supervised contrastive uncertainty learning framework that: (1) treats predictive uncertainty as a representation level supervisory signal rather than solely a post-hoc reliability estimate or filtering mechanism;
(2) structures representation geometry by aligning low uncertainty same class voxel features while preventing ambiguous high uncertainty features from collapsing into confident clusters;
(3) demonstrates significant surface distance (SD) improvement across 6 of 8 anatomical structures ($p<0.001$), with a $22{\times}$ asymmetry between significantly improved and degraded hard cases; and
(4) validates each component's contribution via two ablations: removing uncertainty guidance achieves 5/8 significant SD anatomical structures; replacing the contrastive term with a boundary loss achieves 4/8, confirming uncertainty guided selection as the decisive factor, improving transition zone, while the contrastive mechanism additionally accounts for bladder. To our knowledge, uncertainty has not been explicitly used as a contrastive structuring signal to improve segmentation robustness.


\begin{figure}[h]
    \centering
    \includegraphics[width=1\linewidth]{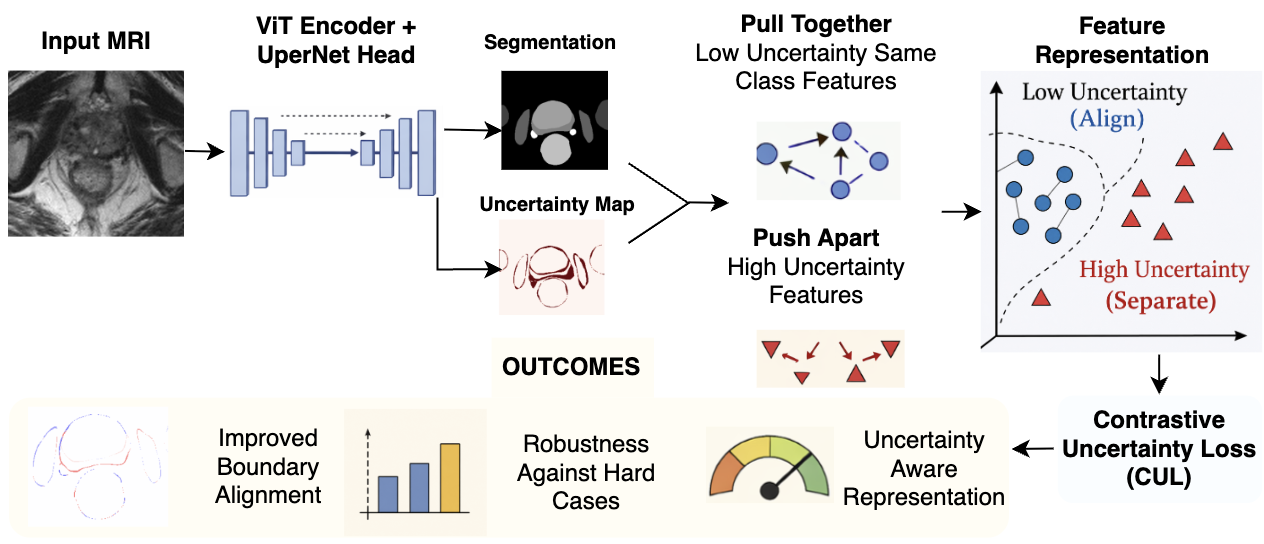}
    \caption{Overview of the proposed Contrastive Uncertainty Learning (CUL) framework. Prostate MRI is processed by a segmentation network to produce voxel wise predictions and an associated uncertainty map. Voxel embeddings are partitioned into low and high uncertainty regions and used to construct contrastive pairs. Low-uncertainty same-class features are pulled together, while high-uncertainty or misclassified features are repelled, improving boundary robustness.}
    \label{fig:placeholder}
\end{figure}
\section{Method}
\subsection{Model Architecture}
For our proposed CUL, we use a 3D model based on a Vision Transformer (ViT) encoder pretrained \cite{wang2026profoundmoderatesizedvisionfoundation} using masked autoencoding, followed by a UperNet head \cite{xiao2018unified} adapted for the segmentation task. For specific formulation of model architecture, let $x \in \mathbb{R}^{D \times H \times W}$ denote the T2W MRI input and $y\in \mathbb{R}^{C \times D \times H \times W}$ the voxel-wise ground-truth masks. The encoder-adapter stack produces multi-scale ($s$) spatial features $\{F^{(s)}\}_{s=1}^{4} = f_{\theta}(x)$ which are decoded using the UperNet decoder into voxel-wise logits $Z = g_{\phi}(\{F^{(s)}\}) \in \mathbb{R}^{C \times D \times H \times W}$. Segmentation probabilities are then defined as $P = \mathrm{softmax}(Z)$. For the baseline segmentation we have a Dice + Cross-Entropy (DiceCE) loss: $\mathcal{L}_{\text{seg}} = \mathcal{L}_{\text{DiceCE}}(Z, y)$

\subsubsection{Predictive Uncertainty Estimation}
For computing our proposed CUL loss, we first compute predictive uncertainty map using Monte Carlo (MC) Dropout in the decoder. Specifically, given $M$(=5) stochastic forward passes, let $P^{(m)}$ be the probability map from pass $m$. The predictive entropy for voxel-wise uncertainty map is then computed as $U = - \sum_{c=1}^{C} \bar{P}_c \log(\bar{P}_c + \epsilon)$
where $\bar{P} = \frac{1}{M}\sum_{m=1}^{M} P^{(m)}$ is predictive probability, $C=8$ is the number of segmentation classes (zones), $P_c$ is the probability for class $c$ and $\epsilon$ is numerical constant for stability. The map is normalised per sample:
$\tilde{U} = \frac{U - \min(U)}{\max(U) - \min(U) + \epsilon}$.

\subsection{Contrastive Uncertainty Learning (CUL)}

\subsubsection{Uncertainty Guided Sampling and Pair Construction:}
Using the uncertainty map, CUL operates on decoder features (before the final classifier), which are projected into a contrastive embedding space. We first define the low-uncertainty mask and prediction-error map:
\begin{align}
M_{\text{low}} = \indicator\!\left(\tilde{U} < \tau\right), 
~E_{\text{map}} = \indicator\!\left(\hat{y} \neq y\right),
\end{align}
where $\tau(=0.5)$ is a fixed uncertainty threshold and $\hat{y}$ is the predicted label.
From these, we define an anchor pool and a negative pool:
\begin{align}
M_{\text{anchor}} = M_{\text{low}} \cdot (1 - E_{\text{map}}), 
~M_{\text{neg}} = \mathds{1}\!\left((1 - M_{\text{low}}) + E_{\text{map}} > 0\right).
\end{align}


\noindent\textbf{Terminology:}
Anchors are voxels ($e_p$) sampled from decoder features using the mask $M_{\text{anchor}}$, corresponding to low-uncertainty and correctly predicted regions. For each anchor, a positive sample is defined as a same-class voxel drawn from the anchor pool, excluding the anchor itself when possible. Negatives ($e_n$) are sampled using $M_{\text{neg}}$, comprising voxels that are either high-uncertainty or misclassified, and are not restricted to different-class locations.


\subsubsection{Uncertainty-Aware Contrastive Loss:}
Projected features (anchors and negatives) are $\ell_2$-normalised, and
for each anchor $e_p^{i}$, one positive $e_{p(i)}^{i}$ is sampled from the same-class confident set, and a set of negatives $\mathcal{N}_i$ is sampled using $M_{\text{neg}}$. The contrastive term is then computed voxel-wise as:
\begin{align}
\mathcal{L}_{\text{CUL-ctr}}^{(i)}
=
-\log
\frac{
\exp(\mathrm{sim}(e_p^{i}, e_{p(i)}^{i})/T)
}{
\exp(\mathrm{sim}(e_p^{i}, e_{p(i)}^{i})/T)
+
\sum_{k \in \mathcal{N}_i}
\exp(\mathrm{sim}(e_p^{i}, e_{n}^k)/T)
},
\end{align}
where $T$ is a learnable temperature parameter. The full training objective is defined as sum of DiceCE and CUL (weighted by learnable parameter $\lambda_{ctr}$):
$\mathcal{L}
=
\mathcal{L}_{\text{seg}}
+ \lambda_{\text{ctr}} \mathcal{L}_{\text{CUL-ctr}}.
$




\section{Experiments and Results}

\textbf{Dataset and Experimental Setup:}
For our experiments, we employ a multi-institute pelvis MR dataset \cite{li2023prototypical}, consisting of 589 T2W 3D MRI input samples of size (D=64, H=256, W=256) with anatomy/zone masks. In total, there are 8 anatomical structures; bladder, bone, obturator internus, central gland, transition zone, rectum, seminal vescicle and neurovascular bundle. The dataset was split into train, validation and test sets (412, 60, 117) and the inputs were preprocessed and augmented using standard preprocessing and augmentation (centre cropping, rotation, flipping, zooming and intensity normalisation). Training used an NVIDIA A40 GPU for 100 epochs with learning rate $1e-4$. The Dice with cross entropy loss was adopted to train our baseline segmentation model using the Adam optimiser. Batch size was set to 4 with 5 Monte Carlo (MC) samples computed for the uncertainty map during training. For performance evaluation, we use Dice, Hausdorff-distance and Surface Distance, capturing the overall overlap as well as worst-case boundary errors and surface errors.  To isolate each component's contribution, we evaluate two ablations under identical training conditions: a class-based variant replacing uncertainty-guided anchor selection with ground-truth class identity only; and a boundary loss variant replacing the contrastive term with a Kervadec-style distance-transform boundary loss. 


\begin{table}[tb]
\centering
\caption{Overall segmentation performance for Baseline, CUL, and the two ablations. SD(mm) is significantly improved which is consistent with CUL's design: predictive uncertainty concentrates at local boundary regions, so improvement manifests in surface-sensitive metrics rather than in global overlap (Dice) or gross deviation extremes (HD95(mm)).}
\label{tab:main_results}
\setlength{\tabcolsep}{4pt}
\renewcommand{\arraystretch}{1.2}

\begin{tabular}{lcccc}
\toprule
\textbf{Metric} & \textbf{Baseline} & \textbf{CUL} & \textbf{Class-Based} & \textbf{Classic BL} \\
\midrule

Dice ($\uparrow$)
& $0.8335 \pm 0.0647$
& $0.8338 \pm 0.0651$
& $0.8331 \pm 0.0637$
& $\mathbf{0.8338 \pm 0.0634}$ \\

HD95(mm) ($\downarrow$)
& $5.9343 \pm 8.1414$
& $\mathbf{5.4130 \pm 4.8829}$
& $5.3736 \pm 4.6558$
& $5.6253 \pm 5.2306$ \\

SD(mm) ($\downarrow$)
& $1.6076 \pm 2.5101$
& $\mathbf{1.4190 \pm 1.8135}$
& $1.4385 \pm 1.8873$
& $1.4641 \pm 1.9188$ \\

\bottomrule
\end{tabular}

\end{table}

\noindent\textbf{Overall Quantitative Performance:}
Table~\ref{tab:main_results} summarises overall segmentation performance across 117 test cases. CUL achieves comparable Dice performance to the baseline (0.8338 vs 0.8335), indicating that uncertainty-aware contrastive learning does not on average degrade global overlap. As CUL specifically focuses on boundary deviations, we observe consistent improvement in boundary sensitive metrics. Mean HD95 decreases from 5.93 mm to 5.41 mm, and mean surface distance (SD) decreases from 1.61 mm to 1.42 mm.

The HD95 standard deviation is notably reduced (8.14 → 4.88 mm), suggesting suppression of extreme outliers, though formal variance testing (Brown-Forsythe $p{=}0.538$) does not confirm this at $\alpha{=}0.05$, likely reflecting limited statistical power at $n=117$; we therefore treat it as a descriptive finding. Paired Wilcoxon signed-rank testing confirms that SD reduction is statistically significant ($p{<}0.0001$), while Dice ($p{<}0.181$) and HD95 ($p{<}0.816$) median shifts are not. Dice is expected to remain unchanged because CUL targets local boundary uncertainty rather than volumetric overlap. Although HD95 is not significant, its mean decreases by 0.49 mm and variance is descriptively reduced, suggesting suppression of extreme outliers despite limited statistical power.

\begin{table}[t]
\centering
\caption{Case-wise performance distribution (n=117). Significant improvement/degradation thresholds: SD~$\pm$0.1~mm, HD95~$\pm$0.5~mm.}
\label{tab:casewise_distribution}
\setlength{\tabcolsep}{4pt}
\renewcommand{\arraystretch}{1.05}
\begin{tabular}{lccc}
\toprule
\textbf{Metric} & \textbf{Improved / Degraded (\%)} & \textbf{Sig+ / Sig- (\%)} & \textbf{Ratio} \\
\midrule

Dice ($\uparrow$) 
& 64 / 53 (54.7 / 45.3) 
& -- 
& -- \\

HD95(mm) ($\downarrow$) 
& 66 / 51 (56.4 / 43.6) 
& 12 / 8 (10.3 / 6.8) 
& 1.5$\times$ \\

SD(mm) ($\downarrow$) 
& 79 / 38 (67.5 / 32.5) 
& 22 / 1 (18.8 / 0.9) 
& 22$\times$ \\

\bottomrule
\end{tabular}
\end{table}

\begin{table}[t]
\centering
\caption{HD95(mm) change stratified by baseline difficulty (median=4.0032).}
\label{tab:hard_case_analysis}
\setlength{\tabcolsep}{6pt}
\renewcommand{\arraystretch}{1.05}
\begin{tabular}{lcc}
\toprule
\textbf{Subset} & \textbf{n} & \textbf{$\Delta$HD95 (mm)} \\
\midrule

Easy ($<$4.0032)  & 58 & $\approx 0$ \\
Hard ($\geq$4.0032) & 59 & \textbf{$-1.1131$} \\

\bottomrule
\end{tabular}
\end{table}

\noindent\textbf{Case-Wise Performance Distribution:}
To characterise model behaviour beyond aggregate statistics, we analyse per-case performance (Table~\ref{tab:casewise_distribution}) across 117 test cases. A case is classified as notably improved or degraded if its mean metric change across all 8 structures exceeds a fixed medically critical absolute threshold (SD: $\pm$0.1 mm; HD95: $\pm$0.5 mm). Dice improves in $54.7\%$ of cases versus $45.3\%$ degraded, indicating only a modest net overlap gain. In contrast, boundary-sensitive metrics show clear asymmetry. HD95 improves in $56.4\%$ of cases (+15 net), while SD improves in $67.5\%$ of cases (+41 net). When considering significant changes, SD exhibits strong directional bias: $18.8\%$ of cases show significant reduction compared to only $0.9\%$ significant increase (22 vs 1 cases). This 22× ratio suggests asymmetric boundary refinement rather than random fluctuation. Figure 2 demonstrates the case where CUL has improved the boundary uncertainty the most (HD95=17.11 vs 4.09) and also the most degraded case (HD95=10.97 vs 12.80).

\noindent\textbf{Hard-Case Robustness Analysis:}
We stratify cases using the median baseline HD95 threshold (4.00) to distinguish easy and hard subsets (Table~\ref{tab:hard_case_analysis}). CUL achieves the largest boundary improvement in hard cases, with a mean HD95 reduction of 1.113 mm, whereas easy cases exhibit only marginal change ($\approx0)$.

\noindent\textbf{Structure-wise Analysis:}
Table~\ref{tab:structure_sd} reports mean surface distance per anatomical structure (paired Wilcoxon,$\alpha{=}0.05$). CUL achieves significant SD improvement in 6/8 structures:bladder, bone, central gland, rectum, transition zone, and neurovascular bundle. Obturator internus and seminal vesicle show no significant change;both have inter-subject variance exceeding three times their mean, suggesting high case to case variability. Dice remains largely unchanged across all structures($|\Delta|{<}0.004$). HD95 reaches significance only for rectum ($p{=}0.004$). Notably, the largest quantitative SD gains appear in bone($-0.16$ mm) and central gland($-0.61$ mm).

\begin{figure}[!ht]
    \centering
    
    \begin{minipage}{1\linewidth}
        \centering
        \includegraphics[width=\linewidth]{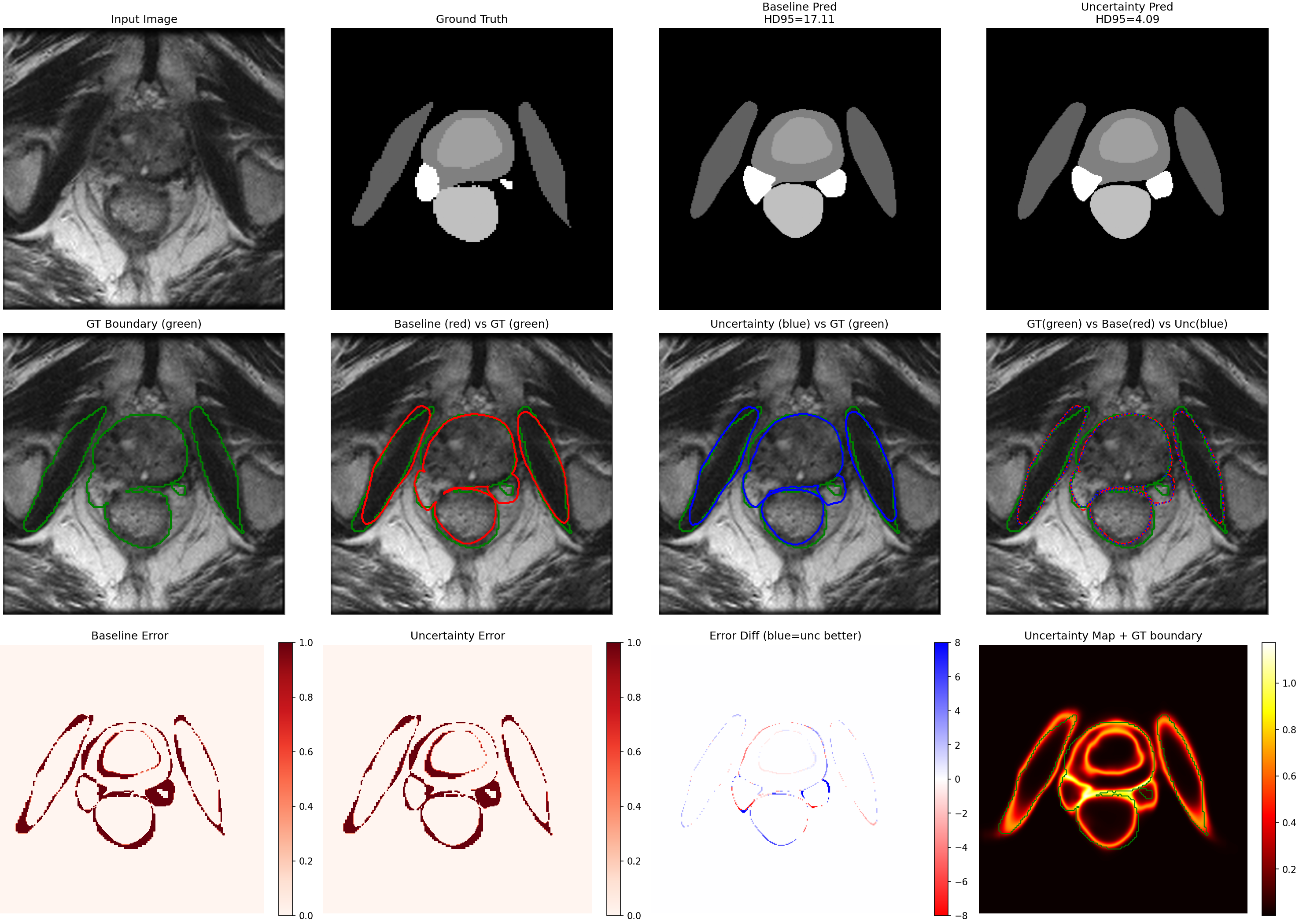}
        \subcaption{Improved case}
        \label{fig:imp}
    \end{minipage}
    \hfill
    \begin{minipage}{1\linewidth}
        \centering
        \includegraphics[width=\linewidth]{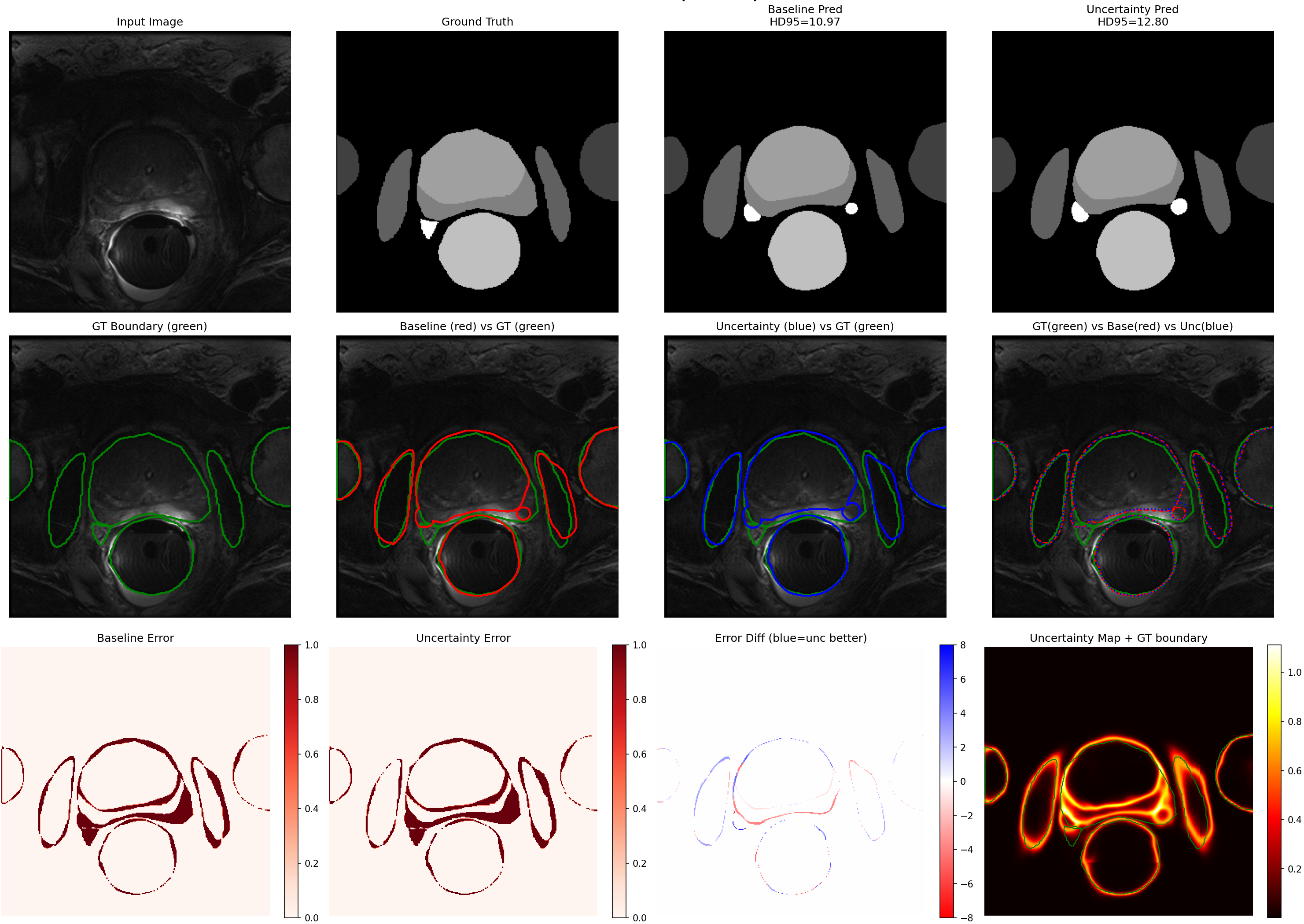}
        \subcaption{Degraded case}
        \label{fig:deg}
    \end{minipage}
    
    \caption{Qualitative comparison between an improved case (top) and a degraded case (bottom). The top row highlights CUL’s role as geometric ``insurance'', where it effectively corrects extreme boundary outliers in cases characterised by high regional ambiguity (e.g. central gland and both neurovascular bundles). CUL organises representations to stabilise boundaries by anchoring confident regions and isolating ambiguity. This visualisation demonstrates the framework's capacity to correct significant spatial deviations while maintaining the integrity of stable predictions.}
    \label{fig:qualitative_comparison}
\end{figure}

\noindent\textbf{Ablation Study:} To isolate the contribution of each CUL component, Table~\ref{tab:structure_sd} also reports the two ablations. The class-based ablation (uncertainty guided selection removed) reaches 5/8 significant structures, failing only on transition zone ($p{=}0.676$) which CUL improves ($p{=}0.002$). This shows uncertainty guidance provides additional benefit beyond class-based contrastive learning. The boundary loss ablation reaches only 4/8, additionally failing to improve bladder ($p{=}0.115$), suggesting geometric loss augmentation cannot replicate uncertainty-guided contrastive structuring. The stepwise pattern, CUL 6/8 > class-based 5/8 > boundary loss 4/8, isolates uncertainty guided sample selection as the decisive factor for transition zone, and the contrastive mechanism as the decisive factor for bladder.


\begin{table*}[ht]
\centering
\caption{Per-structure SD (mm $\downarrow$), n=117, paired Wilcoxon vs.\ Baseline. $^{*}$$p{<}0.05$, $^{**}$$p{<}0.01$, $^{***}$$p{<}0.001$.}
\label{tab:structure_sd}
\setlength{\tabcolsep}{6pt}
\renewcommand{\arraystretch}{1.1}

\begin{tabular}{lcccc}
\toprule
\textbf{Structure} & \textbf{Baseline} & \textbf{CUL} & \textbf{Class-Based} & \textbf{Classic BL} \\
\midrule
Bladder         & $1.17 \pm 1.82$ & $0.85^{***}$ & $0.86^{*}$     & $0.87$          \\
Bone            & $1.20 \pm 1.94$ & $1.03^{***}$ & $1.02^{***}$   & $1.01^{***}$    \\
Obturator Int.  & $1.43 \pm 3.47$ & $1.37$       & $1.30$         & $1.37$          \\
Central Gland   & $2.50 \pm 13.4$ & $1.89^{***}$ & $2.10^{***}$   & $2.02^{**}$     \\
Transition Zone & $1.59 \pm 1.13$ & $1.50^{**}$  & $1.55$         & $1.58$          \\
Rectum          & $1.22 \pm 0.98$ & $1.15^{***}$ & $1.14^{***}$   & $1.13^{***}$    \\
Seminal Vesicle & $1.78 \pm 5.43$ & $1.65$       & $1.62$         & $1.81$          \\
NVB             & $1.97 \pm 2.29$ & $1.93^{***}$ & $1.92^{***}$   & $1.92^{***}$    \\
\midrule
\textbf{Sig.\ structures} & & \textbf{6/8} & \textbf{5/8} & \textbf{4/8} \\
\bottomrule
\end{tabular}
\end{table*}

\section{Discussion and Conclusion}
For our multi-structure MRI segmentation, we propose a supervised contrastive framework for improving geometric robustness, by estimating predictive uncertainty as a representation-level supervisory signal.

Our experiments on a multi-organ prostate MRI segmentation task show that CUL maintains comparable global overlap performance to the baseline, while consistently improving boundary-sensitive metrics. Case-wise and difficulty-stratified analyses further reveal that CUL yields the strongest benefits on hard cases characterised by large boundary deviations. Moreover, comparison with the two ablations clarifies the source of these gains, isolating the uncertainty guided mechanism as the key driver of CUL's improvements in the transition zone.

These improvements have practical implications for radiotherapy planning. In pelvic organ segmentation, surface distance errors of even a few millimetres can translate to incorrect planning target volume (PTV) margins, affecting dose coverage and organ at risk (OAR) sparing. Similar precision requirements are of importance for targeted biopsy and focal therapy. The significant SD reductions, directly targets the boundary precision that affects contour acceptance in clinical workflows. The 22× asymmetry between significant improvements and significant degradations in SD further suggests CUL reduces the tail of outlier errors that typically require manual re-contouring. Future work includes adaptive uncertainty thresholding, additional ablations, and integration with multi-scale, domain-robust training.

\clearpage

\bibliographystyle{IEEEtran}
\bibliography{ref}
\end{document}